\documentclass[conference]{IEEEtran}
\usepackage[letterpaper, left=0.75in, right=0.75in, bottom=0.75in, top=1in]{geometry}

\IEEEoverridecommandlockouts
\usepackage{cite}
\usepackage{amsmath,amssymb,amsfonts}
\usepackage{graphicx}
\usepackage{textcomp}
\usepackage{hyperref}
\usepackage{xcolor}
\usepackage{array}
\usepackage{tabularx}
\usepackage{algpseudocode}
\usepackage{flushend}
\usepackage[ruled,vlined]{algorithm2e}
\def\BibTeX{{\rm B\kern-.05em{\sc i\kern-.025em b}\kern-.08em
    T\kern-.1667em\lower.7ex\hbox{E}\kern-.125emX}}
    \makeatletter
    
\newcommand{\linebreakand}{%
  \end{@IEEEauthorhalign}
  \hfill\mbox{}\par
  \mbox{}\hfill\begin{@IEEEauthorhalign}
}
\usepackage{tikz}
\usetikzlibrary{positioning}
\newdimen\nodeDist 
\newdimen\nodeDistone
\newdimen\nodeDisttwo
\makeatother
\begin{document}

\title{Winning the 3rd Japan Automotive AI Challenge - \\ Autonomous Racing with the Autoware.Auto \\ Open Source Software Stack}

\author{Zirui Zang$^{\dagger}$, Renukanandan Tumu$^{\dagger}$, Johannes Betz$^{\dagger}$, Hongrui Zheng$^{\dagger}$,  and Rahul Mangharam$^{\dagger}$
\thanks{$^{\dagger}$University of Pennsylvania, School of Engineering and Applied Science,
        19106 Philadelphia, PA, USA        
        {\tt\small zzang, nandant, joebetz, hongruiz, rahulm}@seas.upenn.edu}
}

\maketitle

\begin{abstract}
The 3rd Japan Automotive AI Challenge was an international online autonomous racing challenge where 164 teams competed in December 2021. This paper outlines the winning strategy to this competition, and the advantages and challenges of using the Autoware.Auto open source autonomous driving platform for multi-agent racing. Our winning approach includes a lane-switching opponent overtaking strategy, a global raceline optimization, and the integration of various tools from Autoware.Auto including a Model-Predictive Controller. We describe the use of perception, planning and control modules for high-speed racing applications and provide experience-based insights on working with Autoware.Auto. While our approach is a rule-based strategy that is suitable for non-interactive opponents, it provides a good reference and benchmark for learning-enabled approaches.  
\end{abstract}

\begin{IEEEkeywords}
autonomous systems, automobiles, intelligent vehicles, model predictive control, path planning
\end{IEEEkeywords}

\section{Introduction}
\label{sec:introduction}

Autonomous Racing is an efficient research and development setting for safe autonomous vehicles.
In everyday driving, vehicles are designed to be as safe as possible. 
Performance can be difficult to measure in everyday driving maneuvers, such as merging on the highway, or overtaking slower traffic. While performance can be difficult to quantify in these everyday scenarios, the hesitation or aggressiveness of a vehicle in conducting these maneuvers can have a significant impact on safety. Too much hesitation, and the vehicle may interrupt the flow of traffic, becoming a traffic hazard. Too agressive, and the vehicle may cause collisions and reactionary behaviour from other drivers. 

Autonomous racing, on the other hand, penalizes safe but conservative policies so that the need for robust, adaptive strategies is critical. 
Racing adversarial agents magnifies this tension and is an useful setting for testing the limits of safety and performance across the perception, planning and control stack of autonomous vehicles. 
Since the track is known and the sole objective of racing is to minimize laptime without crashing, autonomous racing focuses on achieving this with high speeds, high accelerations, and low reaction times. 
As the opponent’s strategy is secret and cannot be obtained by collecting data before the competition, driving decisions must be made online with high levels of uncertainty in a dynamic and adversarial environment. 

Consequently, autonomous racing \cite{Betz2022}  has become popular over the recent years. Competitions at full scale such as Roborace or the Indy Autonomous Challenge \cite{Wischnewski2022}, as well as at small-scale such as F1TENTH \cite{okelly2020f1tenth}, provide platforms and benchmarks for evaluating autonomous driving software. 
The community's research interests are in two general racing contexts: achieving super-human performance in single-vehicle racing, and performing intelligent overtaking maneuvers against adversarial opponents at the limits of dynamics. 
In the first area, approaches \cite{Heilmeier2019, DalBianco2018,  OKelly2020} usually form a time-optimal, or curvature-optimal problem and solve for a global race line. 
In the second area, approaches such as game theory \cite{sinha2020, Williams2017, Notomista2020} or Reinforcement Learning \cite{schwarting2021, Song2021, Notomista2020} are used to find racing agents able to perform overtakes. 


\subsection{Japan Automotive AI Challenge}
\begin{figure}[h]
  \centering
  \includegraphics[width=0.3\textwidth]{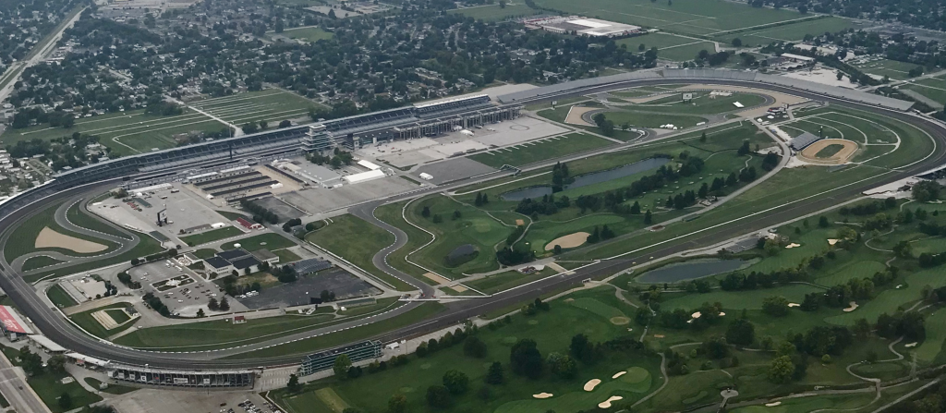}\vspace{0.2cm}
  \includegraphics[width=0.3\textwidth]{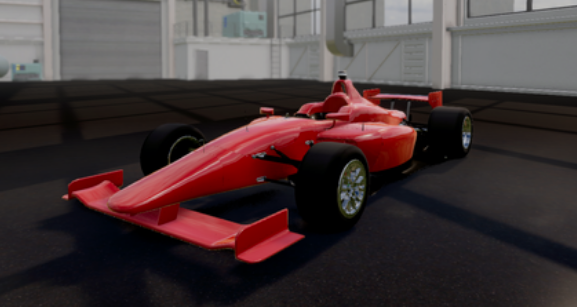}
  \caption{Race Track: The Indianapolis Motor Speedway (top), Race Car: Dallara IL-15 Model (bottom):}
  \label{fig:map_and_car}
  \vspace{-10pt}
\end{figure}

The 3rd Japan Automotive AI Challenge was a worldwide competition hosted by the Society of Automotive Engineers of Japan in December of 2021. The aim of the competition is to help identify and train talent in machine learning applications for future mobility systems. The competition pits teams against the clock in a multi-agent head-to-head autonomous vehicle race. Each competitor takes control of a car, which begins in last place in a lap around the Indianapolis Motor Speedway (shown in Figure \ref{fig:map_and_car}). The competitor is given the ground truth localization of the ego car through the Autoware.Auto GNSS localizer module. There are 5 NPC (Non-Player Character / computer-controlled) cars ahead of the ego, each following a predetermined path. All vehicles in the race are identical to the Dallara IL-15s. The ego must overtake all 5 NPC vehicles, and finish the lap in the shortest time possible. Each collision with NPCs will add a penalty of 5 seconds to the total time. The shortest time around the circuit wins the competition. The event is held entirely in simulation, using the LGSVL simulator~\cite{Guodong2020}, and each team is given the Autoware.Auto software stack to ease the development burden. During evaluation, the submission is scored on the same track, but with the NPCs following a path not known beforehand to the competitors.

This paper has three major contributions:
\begin{enumerate}
    \item We provide a showcase on how to use and leverage and open-source software stack for autonomous driving for an autonomous racing competition.
    \item We explain our approach on creating an opponent detection algorithm, a lane switching overtaking strategy, and the integration of the MPC vehicle controller to be successful in the Japan AI challenge.
    \item We provide insights on the racing strategy and explain both advantages and gaps that need to be filled using the Autoware.Auto open-source autonomous driving software stack.
\end{enumerate}

In the next sections we describe the adaptations necessary to enable Autoware.Auto to be suitable for racing, and the racing strategies we implemented that emerged as the winner. 

\section{Methodology}
\label{sec:methods}

\subsection{Autoware Open Source Stack}
\begin{figure}[h]
  \centering
  \includegraphics[scale=0.33]{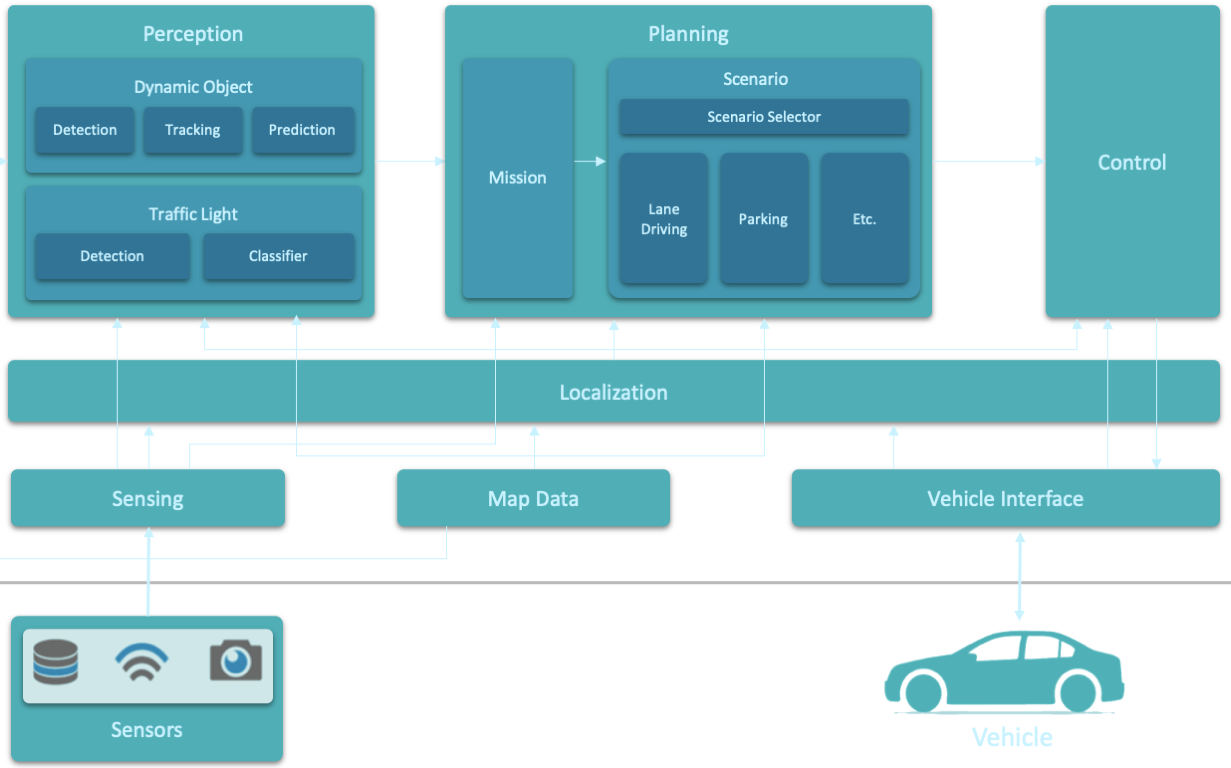}
  \caption{Overview of the Autoware.Auto software stack components}
  \label{fig:autoware}
  \vspace{-10pt}
\end{figure}

Autoware is the world leading open-source autonomous driving software  that combines implementations of perception, planning and control for autonomous vehicle development into one coherent software platform (see Fig. \ref{fig:autoware}). There are two releases of this software, Autoware.AI, which runs on Robot Operating System version 1 (ROS1), and the newer Autoware.Auto, which runs on ROS2 (https://ros.org). Autoware.Auto improves reproducibility and determinism across different system levels, provides modular design, production-grade code and review practices, as well as integrated testing. Previously, Autoware.Auto was used to develop slow-speed autonomous valet parking and cargo delivery services. This competition uses Autoware.Auto for racing with speeds up to 160km/h and for overtaking maneuvers at the limits of the vehicle's dynamics. To support reproducible environments, docker containers with ROS2 Foxy and Autoware.Auto running on Ubuntu 20.04 are distributed to all participants. The interface to the LGSVL simulator, as well as the basic modules needed to get the simulated Dallara IL-15 running, were all provided by Autoware.Auto.

\begin{figure}[!b]
\vspace{-10pt}
  \centering
  \includegraphics[scale=0.170]{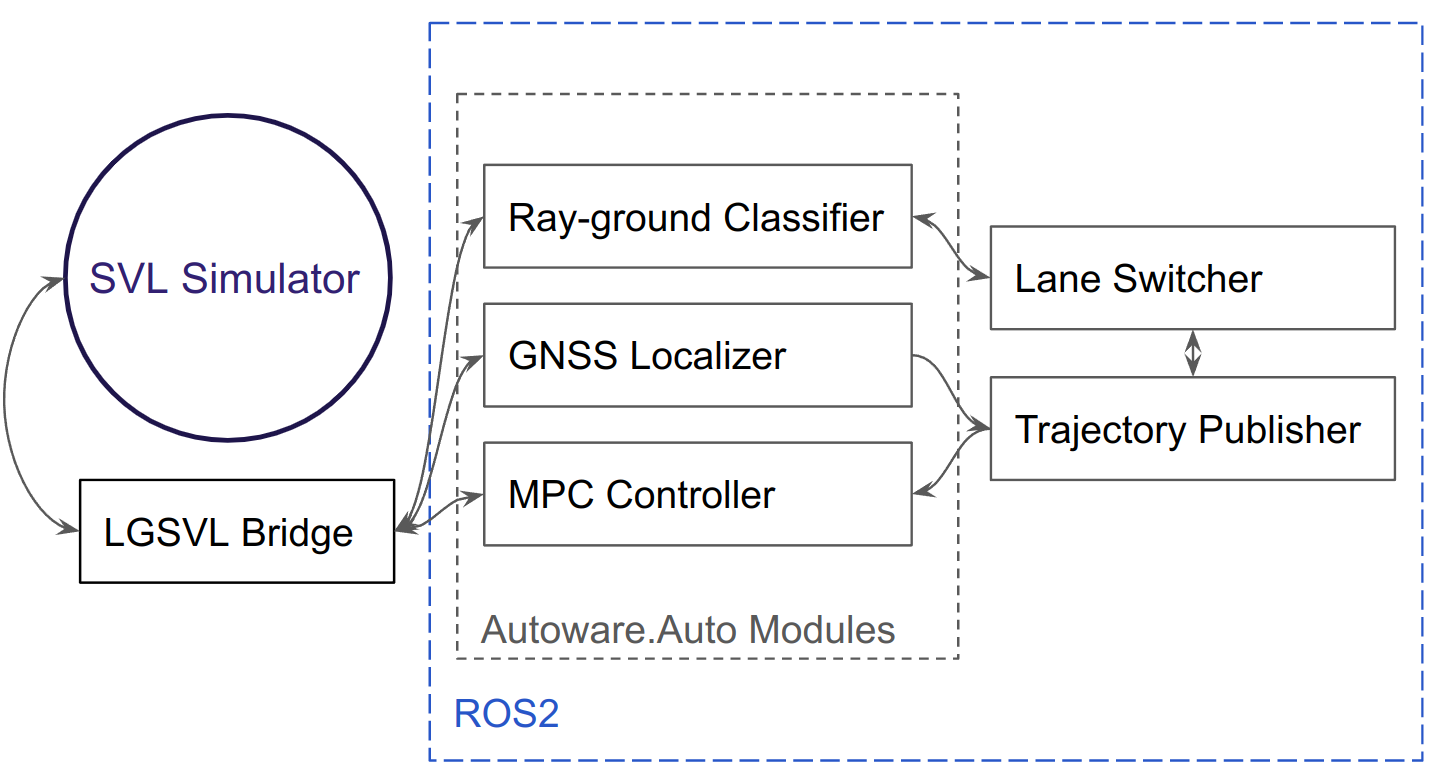}
  \caption{Code Structure for the Race: Our code is integrated with Autoware.Auto modules within ROS2 and communicates with the SVL Simulation through a bridge.}
  \label{fig:structure}
\end{figure}

\figurename{\ref{fig:structure}} shows the code structure of our implementation and enhancements to Autoware.Auto. Our ROS nodes communicate with Autoware.Auto nodes within the ROS2 environment, which is connected with the LGSVL simulation through a LGSVL bridge.

\subsection{Opponent Detection}

The ego car is equipped with a single front-mounted 3D LiDAR to perform obstacle avoidance. Raw LiDAR data is first filtered by the ray-ground filter within the Autoware.Auto framework, which uses the projection of rays to filter out LiDAR points reflected by the ground. The LiDAR data is then further cropped by 3D rectangular boundaries set in the ego vehicle’s frame. In the X axis, we have set a limit from -10 to 100 meters. The -10 meter look-back distance is to detect obstacles in the left and right blind spots of the ego vehicle. In the Y axis, the cropping is based on the current lane the ego vehicle resides in, as we need to filter out the detection from track boundary panels and walls. In the Z axis, we crop the data between -0.5 to 0.9 meters to prevent ground detection points from the banking of the track while the ego car is turning. This is because the ray-ground filter will let some ground points pass through if when the slope of the ground plane is high. We directly use the filtered point cloud data in representing obstacles, which will be later classified into different lanes.

In contrast to our simple approach, the obstacle avoidance pipeline provided by Autoware.Auto is to first combine filtered points into obstacle objects using euclidean clustering, which groups points into clusters if they can be connected by other points within a threshold distance. The obstacle objects will then be examined with the planned trajectory for a potential collision. However, we saw a greater than 700~ms latency between the publishing of the filtered LiDAR point cloud and the publishing of the euclidean clustering results on our machine. This high latency made control of the vehicle unreliable, unable to catch overtaking opportunities, and prone to crashes. Since the LiDAR signal rate is 10~Hz, our computation time should be less than 100~ms to make the best use of the LiDAR information. Therefore, we designed a simple obstacle detection method for this race.

\subsection{Overtaking Strategy using Lane Switching}

\begin{figure}[t]
  \centering
  \includegraphics[scale=0.21]{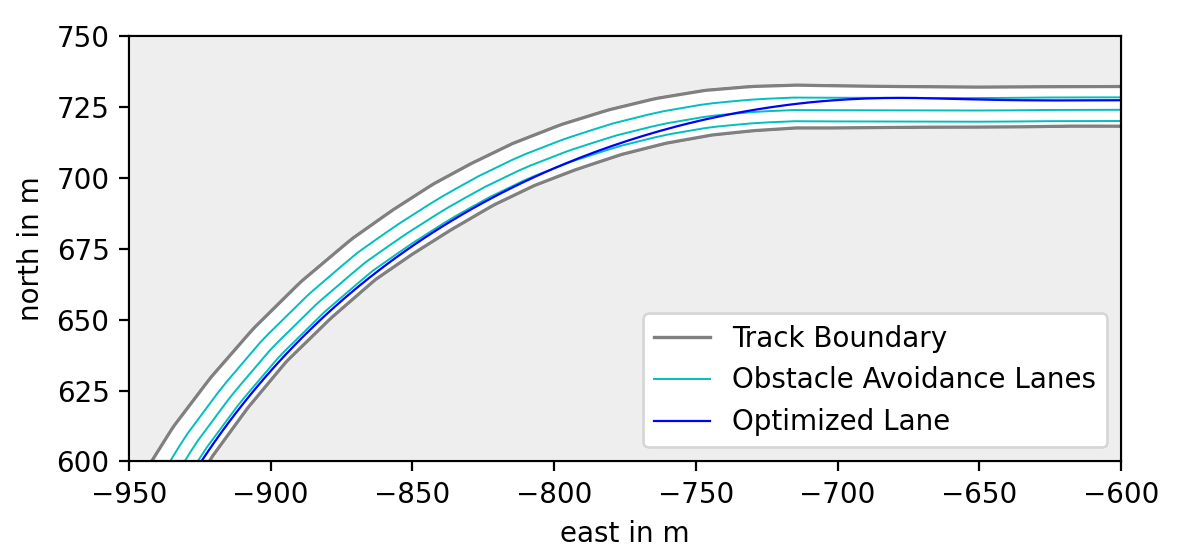}
  \vspace{-20pt}
  \caption{Inner, Center, Outer and Optimized Lanes: The optimized lane will cut across the other three lanes when entering a turn.}
  \label{fig:lanes}
  \vspace{-10pt}
\end{figure}

In this race, the trajectories of the NPCs are based on the inner, center and outer lanes on the racing track. In order to overtake these NPCs with quick and safe maneuvers, we prepared three lane-based racelines and a switching logic. We also prepared a globally optimized raceline, which the ego will switch to if it does not detect any nearby opponent. \cite{Heilmeier2019} \figurename{\ref{fig:lanes}} shows all four lanes when entering curve.

\textbf{NPC Overtaking} - Since the ego vehicle is always positioned behind all NPCs and in the outer lane when each race is initiated, it must overtake all five NPCs to reach the front. We prepared the inner, center and outer lanes by computing equispaced offsets from the centerline of the track, which is provided by the organizers. 

The filtered LiDAR point cloud is then classified into the lanes by calculating the euclidean distance from each LiDAR point to the nearest point on the centerline of each lane. Separate sparse centerlines of each lane are prepared to reduce computation. The number of LiDAR points within each lane are simply counted to give lane occupancy values $l \in [l_0, l_1, l_2]$. The lane occupancy values from the previous detection are also recorded as $l^{pre} \in [l^{pre}_0, l^{pre}_1, l^{pre}_2]$. Two thresholds $\theta_o, \theta_e$ are defined to determine whether the lane is occupied or empty. A lane is marked as occupied only if its lane occupancy value is greater than $\theta_o$, and marked as empty only if smaller than $\theta_e$. The use of two values seems redundant but it allows us to adjust the amount of empty space ahead separately for switching out and into a particular lane. The $\theta_e$ we use is 3 times smaller than $\theta_o$ to give more readiness in the target lane.

The lane switch logic is as follows: The ego vehicle will switch to target lane $t$ from current lane $s$, if $l_s > \theta_o$ and $l_t<\theta_e,~l^{pre}_t < \theta_e$. Else, the ego vehicle will brake and look for further opportunities to switch.

When the current lane is the center, the ego vehicle has both left and right lanes to choose from, otherwise, it will prefer to choose to switch to the center lane. This is because LiDAR detection can be unreliable on the farthest lanes if the vehicle is not in the center, especially around corners where ground filtering is challenging. However, if it must switch across the center lane to a further lane, e.g. left to right directly, then it needs to have  $l_{center} < \theta_o$. After each lane switch is initiated, a pause flag will be on for 10 seconds to prevent additional lane switch signals for the vehicle to settle in the new lane.

\textbf{Globally Optimized Raceline} - The globally optimized raceline is prepared using the minimum curvature method \cite{Christ2019}. Compared to the centerline, an optimized raceline will try to reduce curvature round turnings to allow minimum speed loss while applying less steering. This involves cutting from the outer lane to the inner lane when entering a turn and back to the outer lane when exiting a turn. Maximizing our stay on the optimized lane while effectively avoiding the NPCs has direct impacts on the average speed of our vehicle. Our strategy is that the ego vehicle will try to switch to the optimized lane when we have 5 consecutive detections where all $l_0, l_1, l_2 < \theta_e$. To avoid obstacles while on the optimized lane, we keep track of the effective lane the ego vehicle is currently in and follow a similar lane switching logic as above.

\subsection{Vehicle Control}

Once we have the current or target lane selection, a trajectory publisher will publish correct waypoints based on current localization. The trajectory publisher reads in prepared waypoint files and segments out the correct waypoints from each of the lanes. 

The waypoint data are subscribed to by the Model-Predictive Controller (MPC) module in the Autoware.Auto framework. This MPC module is based on the ACADO toolkit \cite{Houska2011a}. The MPC problem is specified as a Quadratic Problem (QP), with constraints on the acceleration, steering, and speed. A kinematic bicycle model is used for the vehicle. The MPC objective is to minimize the pose error between the vehicle and the given reference trajectory over the planning horizon. The module offers a choice of three solvers: an unconstrained QP solver which uses an eigen-decomposition, a second unconstrained QP solver which uses a faster approximation of the eigen-decomposition, and the qpOASES solver\cite{ferreau_qpoases_2014, tier_iv_mpc_2022}. The method we used was the unconstrained solver with a faster approximation.

Using a changing reference trajectory posed challenges when tuning the MPC, as we do not explicitly create splines for the transition between lanes, instead, we just provide waypoints from the selected lane and let the MPC optimize a control sequence for the lane switch. The default tuning of the Autoware.auto MPC controller place heavy weights on positional and heading errors, which was meant to increase control precision in low-speed parking movements. At high speed, this leads to heading wobbles and large maneuvers during lane transitions which sometimes cause loss of traction. Therefore, we tuned down about 25\% of the weights for position and heading and tuned up weights for longitudinal velocity by 20 times to prioritize maintaining speed over trajectory precision.

\section{Results}
\label{sec:Results}

In the final evaluation with unknown NPCs, our method is able to complete the race without opponent contact in 99.5784~s, 0.32~s faster the 2nd place, with an average speed of 41.33~m/s.

Compared to the long latency of the euclidean clustering method, our simple lane-based obstacle avoidance algorithm takes an average processing time of 20~ms. This helps us react to LiDAR data as quickly as possible. With the peak speed of more than 47~m/s, 20~ms of processing time translate to roughly 0.94~m in distance where the vehicle cannot perform any updated maneuver. As we can see, in a high-speed racing scenario such as this competition, short processing time can give rewarding advantages.

\begin{figure}[h]
  \centering
  \includegraphics[scale=0.25]{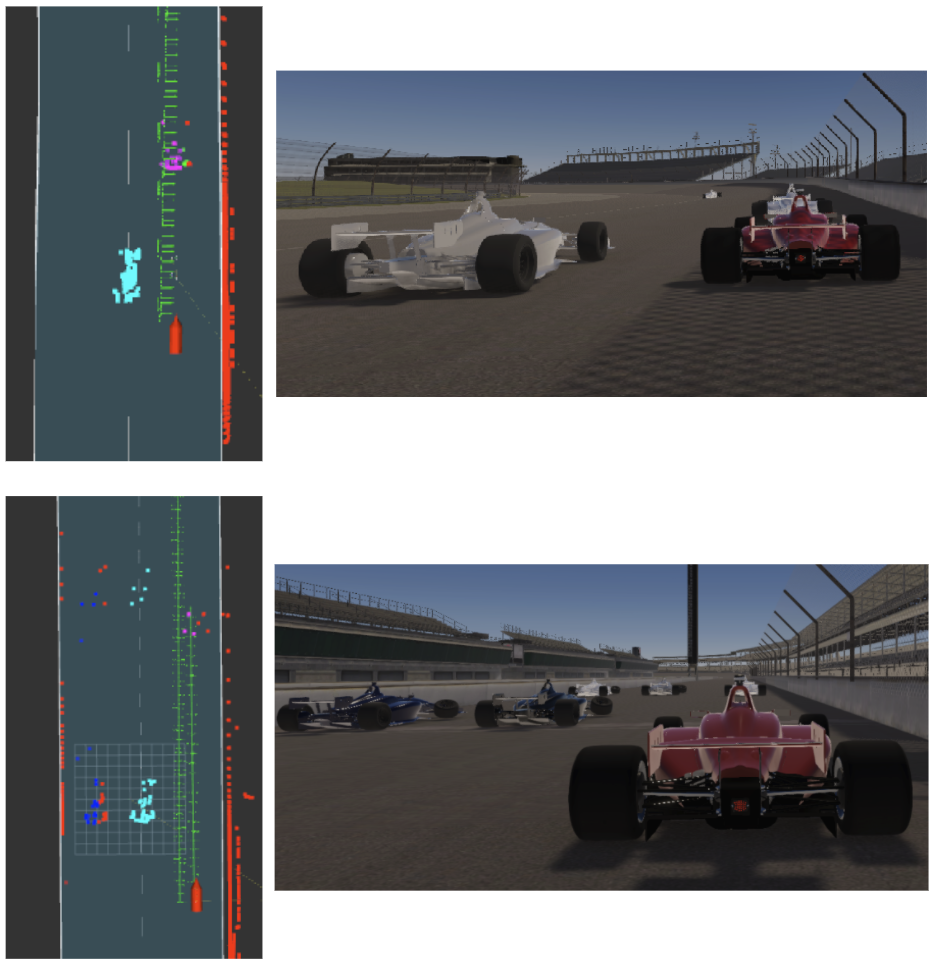}
  \caption{LiDAR Point Classification Examples: Visualization Plots (left), Corresponding Simulation Screenshots (right). Points with different colors belong to different lanes.}
  \label{fig:point_classification}
\end{figure}

\begin{figure*}[h]
  \centering
  \includegraphics[width=\textwidth]{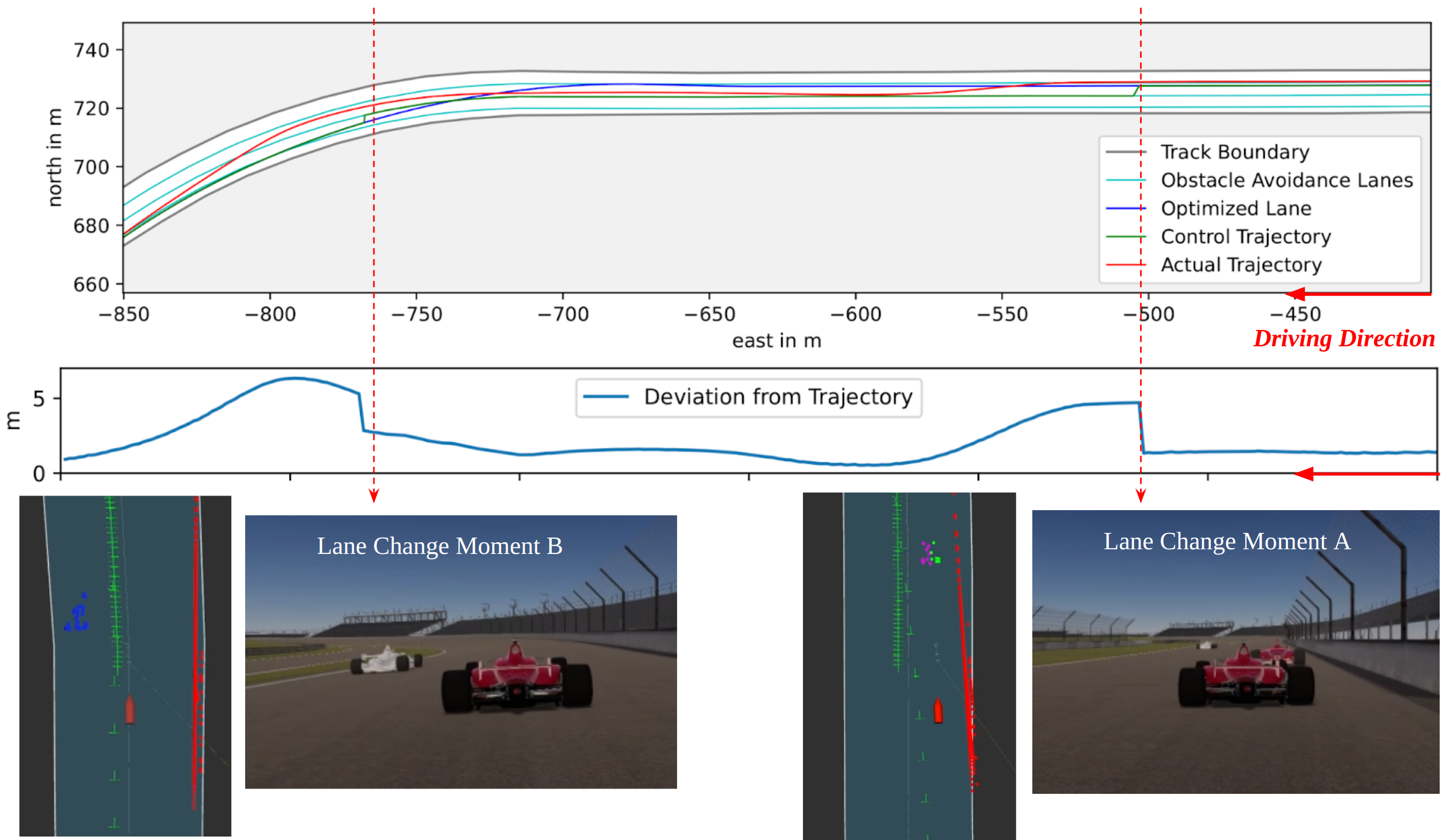}
  \caption{Example Lane Switching Moments: Plot of Lanes and Trajectories (top), Plot of Control Error (center), Simulation and RVIZ Screenshots (bottom): In moment A, the ego switch to the center lane to overtake the front opponent. In moment B, the ego car waits until the inner lane is clear to switch to the globally optimized lane. The driving direction is from left to right.}
  \label{fig:lane_switch}
  \vspace{-10pt}
\end{figure*}

Two examples of in-race LiDAR point cloud classifications are shown in \figurename{\ref{fig:point_classification}} with visualization plots on the left and a live simulation view on the right. On the bottom is the beginning position of the race. In the visualization plot, we can see points colored in blue, cyan and magenta belong to different lanes, and points that are filtered out are colored in red. On the top is a scenario where the ego car is blocked by a leading car in the outer lane but unable to overtake due to the middle lane being occupied as well. In this case, the car will slow down to avoid a collision penalty and also make space for other overtaking opportunities.

In \figurename{\ref{fig:lane_switch}}, we present two consecutive overtake examples that happened in a training session. Cars are driving from the right side and entering a left-turning curve. At the trajectory plot at the top, we can see two shifts in the green curve which is the control trajectory published by the trajectory publisher. In lane change moment A, the ego car switched from the optimized lane to the middle lane to perform an overtake of an opponent vehicle. After the overtake, the ego car should switch back to the optimized lane, which will cut into the inner lane to minimize curvature. However, at this time, in moment B, the inner lane was occupied. The ego vehicle correctly detected the presence of the occupying opponent and postponed the lane change until the inner lane was cleared. We can also see the actual trajectory plotted as the red curve and the control error plotted in the middle graph, which shows smooth trajectory was calculated with the Autoware.Auto MPC controller despite two sharp lane switches. We have recorded a \href{https://youtu.be/awtK3_frer8}{\color{blue}video} from training session to demonstrate lane switching overtakes.

During our experiments, a frequent fail case of this method that we observed is when the target lane that the ego is switching to can be blocked by another car that is a short distance ahead. This is usually because the lidar detection for the target lane is partially blocked by the opponent car in front of the ego due to its close up distance. For example, at lane change moment A in \figurename{\ref{fig:lane_switch}}, the ego is about to switch to the center lane, but there is a blind zone in the center lane blocked by the front car. If this happens, once the ego switches to the center lane, it may have to brake hard to avoid collision. A potential solutions to this scenario will be to track the dynamic of the front car while they are visible.

\section{Discussion}
\label{sec:Discussion}

The Open Source Stack provided by Autoware provides a great advantage in getting started with full-stack autonomous vehicle development. In this race, it allowed us to focus on the areas of perception, planning and control that mattered the most for the race. The LiDAR pre-filtering and MPC control worked well. It also provided easy to use interfaces between our code and the simulator.

While access to ready-made algorithms expedited development, there were a few areas in which the Autoware stack was not sufficient for autonomous racing. Many of the existing modules, including the global and behavior planner, are specifically designed for low-speed applications like valet parking. The default logic of collision avoidance is to stop, which is not favored and can be dangerous in racing. Instead, race cars need active planning to avoid obstacles and search for opportunities for overtaking. The software stack lacks hardware accelerated implementations of algorithms. For example, the refresh frequency of the euclidean clustering for LiDAR scans can be greatly improved with a GPU implementation. While lower frequencies may be sufficient for low-speed driving scenarios, they are not suitable for higher-speed scenarios like those we encountered in the races.

Racing other opponents presents a challenge, namely that of anticipating the next moves of the opponents. Some related work has made significant progress here, using game theoretic approaches to predict opponent trajectories. Still other approaches use data driven methods to identify moving obstacles, and their possible trajectories. Our algorithm would perform better if we were able to anticipate the future poses of our opponents, and use that information to execute smoother and less reactive lane changes.

The heavily structured nature of the race with non-interactive opponents allowed our largely reactive algorithm to be successful. 
These predefined NPC trajectories meant we could treat our opponents as simple moving obstacles, which made our lane switching approach highly effective. Using an optimized raceline provided us with the lap time benefit needed to win the challenge.

\section{Related Work}
\label{sec:related_work}
Here we cover recent work in autonomous racing and overtaking. While much of the related work in planning racelines is non-reactive, work on overtaking does assume dynamic opponents.

\subsection{Hierarchical Planning}
We view the challenge as a hierarchical planning problem.
Globally, we aim to find an optimized reference raceline that can achieve minimal lap time if tracked flawlessly.
Locally, we aim to find a local plan that deviates from the global plan in the presence of an obstacle, or dynamic opponents, and tracks the global plan accurately otherwise.
In the field of AR, there are numerous efforts on addressing this problem. In the following discussion, we compare and contrast different approaches both in global planning and local planning.

\textbf{Global Planning} - In Global Planning, we can roughly categorize approaches by the objective function used.
First, lap times are used as the optimization objective. In \cite{Bevilacqua2017, Quadflieg2011, OKelly2020}, Evolutionary Algorithm (EA) based optimization is used. Each approach parameterizes the search space differently, and uses different EAs while maintaining the same overall goal. In \cite{Metz1989, Kelly2010, Rucco2015, Theodosis2012, Pagot2020, Vazquez2020, Herrmann2019, Herrmann2020, Herrmann2020_2, Lovato2021, Christ2019}, an Optimal Control Problem (OCP) is formed, with different constraints in vehicle dynamics and geometry limits to minimize lap times. 
Second, certain geometric properties of the final reference raceline have also been parameterized as the optimization objective. In \cite{Braghin2008, Cardamone2010, Heilmeier2019}, an optimization is formed to minimize the maximum, or total curvature of the resulting reference raceline. Third, some approaches also aim to mimic the driving behavior of a race car driver geometrically. For example, \cite{Theodosis2011, Kuhn2017, Lovato2021}.
Our global plan is generated following the approach in Christ et al. \cite{Christ2019}, which is a minimum time approach.

\textbf{Local Planning} - In Local Planning, we can group different methods by their overall strategy.
First, modifying the global plan via optimization by including obstacles into the constraint, or the objective of the problem. \cite{Anderson2016, Kapania2016, Funke2017, Subosits2019, Alcal2020_2, ICRAworkshop_08, Brudigam2021} all falls into this category. These approaches either try to balance two competing objectives in achieving higher speeds and being reactive to obstacles, or perform mode switches to change the weighting on the objective.
Second, sampling multiple dynamically feasible trajectories (motion primitives), and selected via cost minimization. \cite{Williams2016, Liniger2014, Liniger2015, Stahl2019_2, sinha2020} all generates motion primitives with a series of randomly perturbed control sequences, series of fixed inputs, or dynamically feasible splines with different goal points. A cost is usually assigned to each of the primitives, and the one with lowest  overall cost is chosen. Our winning strategy for local planning can be categorized into this group. In our approach, instead of creating locally feasible primitives, we create global primitives, and create the motion plans for switching between these primitives with an MPC.
Lastly, sampling in the free space around the obstacle and construct an obstacle-free path in the explored space. This type of approach is akin to a traditional motion planning problem. \cite{Jeong2013, Arslan2017, Feraco2020, Bulsara2020} all uses variants of RRT to find collision free paths in the free space for racing.

\subsection{Learning-based Planning}
Alternatively, the problem could be considered holistically. Many approaches provide end-to-end, or partially end-to-end solutions to the planning problem. Instead of finding an optimal global raceline, the lap time objective is formulated as part of the reward function. 

On one hand, Reinforcement Learning (RL) is used to train an agent in an adversarial environment. DeepRacing \cite{weiss2020,Weiss2020_1, ICRAworkshop_05} provides solutions on three levels: pixel to control, pixel to waypoints, and pixel to curves. \cite{Perot2017, Jaritz2018} uses A3C to train racing agents. \cite{deBruin2018} learns state representation and uses Q-learning to provide generalizable racing policies. \cite{Remonda2019, Niu2020} uses DDPG to train racing policies. SAC is also widely used \cite{chisari2021, Guckiran2019, Song2021, Fuchs2021}. \cite{schwarting2021, Brunnbauer2021} first learns a latent representation of the world and learns through self-play.

On the other hand, Game Theoretic methods usually abstract the planning problem into a sequential game and tries to find the best response, or minimize regret. \cite{Williams2017, Notomista2020, wang2020, Wang2019, Wang2021} uses Best Response or Iterated Best Response to find best action at each step, or seek an modified Nash equilibrium. \cite{sinha2020} builds an opponent prototype library and uses EXP3 to identify the opponent online. \cite{Liniger2020} plays a sequence of coupled games in a receding horizon fashion. And finally \cite{Schwarting2021_2} uses local iterative DP to plan in Belief Space.




\section{Conclusion and Future Work}
\label{sec:Conclusion}

We have shown the racing strategy used to create the winning entry in the Third Japan Automotive AI Challenge. Leveraging the Autoware.Auto open source autonomous driving software stack allowed us to create the perception, planning and control methods for autonomous racing in just a few weeks. We demonstrated the usability and robustness of the Autoware.Auto modules as well as potential improvements and changes needed for high-speed, high-risk driving scenarios in autonomous racing. Beyond this race, we will continue to work with and develop for projects such as Autoware to support open-source developments in the autonomous driving industry.

\bibliographystyle{IEEEtran}
\bibliography{IEEEabrv,main.bib}

\end{document}